\pdfoutput=1

\documentclass[11pt]{article}

\usepackage[final]{acl}

\usepackage{times}
\usepackage{xcolor}
\usepackage{colortbl}
\usepackage{latexsym}
\usepackage[shortlabels]{enumitem}
\usepackage{graphicx}
\usepackage{amssymb}
\usepackage{amsmath}
\usepackage{adjustbox}
\usepackage{float}
\usepackage{pgfplots}
\usepackage{tikz}
\usepackage{multicol}
\usepackage{multirow}
\usepackage{pgfplotstable}
\usepackage{csquotes}
\usetikzlibrary{patterns}
\usepackage{booktabs}
\usepackage{tabularx}

\definecolor{lightgreen}{RGB}{153, 204, 153}

\usepackage[T1]{fontenc}

\usepackage[utf8]{inputenc}

\usepackage{microtype}

\usepackage{inconsolata}

\title{Trace-of-Thought Prompting: Investigating Prompt-Based Knowledge Distillation Through Question Decomposition}

\author{Tyler McDonald \\
  Department of Computer Science \\
  Brock University \\
  St. Catharines, Ontario \\
  \texttt{tmcdonald3@brocku.ca} \\\And
  Ali Emami \\
  Department of Computer Science \\
  Brock University \\
  St. Catharines, Ontario \\
  \texttt{aemami@brocku.ca}
}
\begin{document}
\maketitle
\begin{figure}[t]
    \centering
    \includegraphics[height=12.5cm, width=\columnwidth, keepaspectratio]{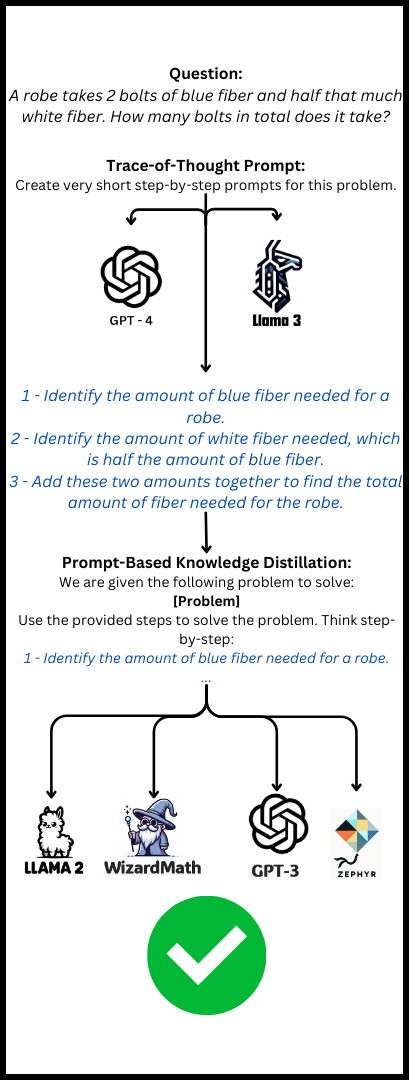}
    \caption{A Visual Depiction of our Trace-of-Thought prompting strategy on a GSM8K problem instance \citep{cobbe2021training}.}
    \label{fig:coverfig}
\end{figure}
\begin{abstract}
Knowledge distillation allows smaller neural networks to emulate the performance of larger, teacher models with reduced computational demands. Traditional methods for Large Language Models (LLMs) often necessitate extensive fine-tuning, which limits their accessibility. To address this, we introduce Trace-of-Thought Prompting, a novel framework designed to distill critical reasoning capabilities from high-resource teacher models (over 8 billion parameters) to low-resource student models (up to 8 billion parameters). This approach leverages problem decomposition to enhance interpretability and facilitate human-in-the-loop interventions. Empirical evaluations on the GSM8K and MATH datasets show that student models achieve accuracy gains of up to 113\% on GSM8K and 21\% on MATH, with significant improvements particularly notable in smaller models like Llama 2 and Zephyr. Our results suggest a promising pathway for open-source, low-resource models to eventually serve both as both students and teachers, potentially reducing our reliance on high-resource, proprietary models.

\end{abstract}

\section{Introduction}
Knowledge distillation, as initially proposed by \citet{hinton2015distilling}, involves leveraging the outputs of larger neural networks as soft targets to train smaller, more efficient networks. This method, primarily applied to tasks like MNIST \cite{lecun1998gradient} in computer vision, uses computationally heavy teacher models to facilitate equivalent reasoning capacities in smaller models, substantially reducing computational demands on the user. As the popularity of Large Language Models (LLMs) has surged, adaptations of this technique have been explored, particularly through fine-tuning based on the outputs of these large models. However, these adaptations often introduce a significant computational overhead and necessitate a deep understanding of machine learning, limiting their accessibility for average consumers \citep{xu2024survey, gu2024minillm, liu2024evolving, zhong2024panda}.

Concurrently, the rapid development of LLMs has been paralleled by innovations in prompt engineering—the strategic design of prompts to enhance reasoning and explore various problem-solving pathways \citep{sahoo2024systematic, chen2024unleashing}. Methods such as Chain-of-Thought Prompting and Self-Consistency have demonstrated the potential of LLMs to engage in complex reasoning and provide novel solutions to challenging problems \citep{wei2023chainofthought, wang2023selfconsistency, yao2023tree, wang2023planandsolve}. Nevertheless, these approaches typically operate within a single contextual framework and rely heavily on the innate reasoning capabilities of models, often failing when applied to smaller, open-source variants \citep{touvron2023llama, tunstall2023zephyr, xu2023wizardlm}. This suggests a critical need for a more adaptable and scalable approach to knowledge distillation that can leverage the advances in prompt engineering for broader accessibility and effectiveness.

In response to this need, we explore the intersection of prompt engineering and knowledge distillation through a novel concept we term \textit{prompt-based knowledge distillation}. This approach utilizes in-context learning (ICL) to emulate traditional distillation processes within the accessible framework of LLM prompting, mirroring the cognitive process of a student learning from a teacher \citep{brown2020language}. To implement this concept, we introduce \textit{Trace-of-Thought Prompting}, a technique that decomposes complex arithmetic reasoning problems into manageable steps, facilitating the distillation of critical reasoning skills from high-resource models to their low-resource counterparts (see Figure \ref{fig:coverfig}). This strategy not only improves the performance of low-resource models but also demonstrates their potential to serve as effective teachers themselves.

Our contributions to this novel extension of knowledge distillation are threefold:
\begin{enumerate}
\item We propose \textit{Trace-of-Thought Prompting}, a novel framework for prompt-based knowledge distillation. This approach allows knowledge transfer from high-resource models (greater than 8 billion parameters) to low-resource models (up to 8 billion parameters) through structured problem decomposition. 

\item We demonstrate significant performance enhancements across two complex arithmetic reasoning datasets. By applying Trace-of-Thought Prompting, we improve the performance of low-resource models on the GSM8K dataset by 113\% and on the MATH dataset by XYZ. Our results also illustrate the effectiveness of low-resource models, like Llama 2 and Zephyr, in achieving performance gains that make them viable alternatives to their high-resource counterparts.

\item Our extended analyses demonstrate that Trace-of-Thought Prompting not only enhances quantitative performance metrics but also improves the transparency of the problem-solving process. This transparency allows for more effective human-in-the-loop interventions, where incorrect or suboptimal reasoning paths generated by the models can be identified and corrected before execution. 
\end{enumerate}

\section{Related Work}
\textbf{Decomposed reasoning.} Traditional question decomposition methods, including Plan \& Solve Prompting and Progressive Hint Prompting, engage in single-context question decomposition, integrating a planning stage followed by an execution phase \citep{wang2023planandsolve, press2023measuring, sun2023pearl}. More sophisticated recursive techniques, such as Least-to-Most Prompting, sequentially append results to enhance the context for subsequent prompts \citep{zhou2023leasttomost, dua2022successive, khot2023decomposed, zheng2023progressivehint}. These methodologies, however, face significant challenges: single-context systems fail to effectively leverage multiple models simultaneously, limiting flexibility and adaptability; recursive techniques, while intricate, hold the potential to lead to extended input sequences and excessive computational demands by virtue of their repetitive nature \citep{guo2024embodied, mohtashami2024social, juneja2024textttlmtexttt2}. Our Trace-of-Thought Prompting addresses these issues by facilitating dynamic, multi-model cooperation without the need for expansive input chains, streamlining the reasoning process across varied contexts.

\textbf{Open-source language modeling.} The rise of open-source models like WizardLM, Zephyr, and Llama has democratized access to language model customization and deployment \citep{xu2023wizardlm, touvron2023llama, tunstall2023zephyr, gunasekar2023textbooks, gemmateam2024gemma}. Despite their accessibility, the teams behind these models report frequent deficiencies in complex reasoning tasks in small variants, underscoring a persistent correlation between model size and reasoning capabilities \citep{agrawal2024llms, chen2024selfplay, zhang2024small}. Trace-of-Thought Prompting enhances these models' performance by distilling complex reasoning from larger models into manageable steps, effectively bridging the gap in reasoning prowess without extensive hardware demands.

\textbf{Tandem and Socratic reasoning.} The exploration of collaborative problem-solving in model suites, such as Socratic Chain-of-Thought and Socratic Questioning, introduces novel ways to utilize multiple models in a cohesive manner \citep{shridhar2023distilling, qi2023art, chang2024socrasynth, zeng2022socratic, goel2024socratic}. However, these approaches encounter difficulties with managing large context sizes and reliance on fine-tuning \citep{li2024longcontext, wang2024adapting}. Our work contributes to this area by implementing a structured approach that minimizes token bloat and fine-tuning dependency, offering a more efficient and scalable solution for collaborative reasoning within LLM environments.

\begin{figure*}
    \centering
    \includegraphics[height=7cm, width=\textwidth, keepaspectratio]{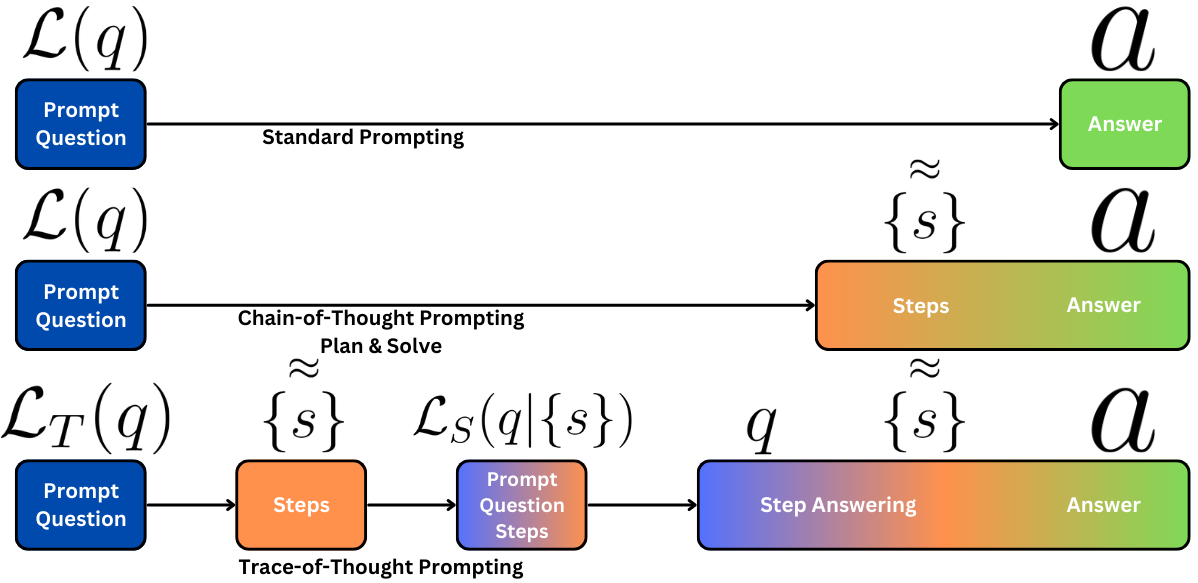}
    \centering
    \caption{Visual depiction of the methods employed during experimentation. Trace-of-Thought provides a novel decomposition framework in a linear manner.}
    \label{fig:mainfig}
\end{figure*}

\begin{table*}[t]
\centering
\small

\begin{tabular}{@{}lp{10cm}@{}} 
\toprule 
\textbf{Prompt Type} & \textbf{Template} \\
\midrule 
Standard & ``<question>.'' \\
Chain-of-Thought & ``<question>. \textit{Think step-by-step.}'' \\
Trace-of-Thought - Delegation & ``\textit{Create very short step-by-step prompts for the following problem:} <question>. \textit{Format as a list. Do not solve the problem.}'' \\
Trace-of-Thought - Solution & ``\textit{We are given the following problem:} <question>. \textit{Use the following steps to solve the problem:} <steps>.'' \\
\bottomrule 
\end{tabular}
\caption{Prompting templates used in experimental evaluation.}
\label{tab:prompt_templates}
\end{table*}

\section{Prompt-Based Knowledge Distillation}

Traditional knowledge distillation, as originally proposed by \citet{hinton2015distilling}, involves fine-tuning smaller neural networks on the soft outputs (logits) of larger, teacher networks. This transfer learning method enhances the smaller model's performance to emulate its larger counterpart, albeit with significantly reduced computational overhead. Despite its effectiveness, traditional knowledge distillation is resource-intensive, necessitating extensive computational efforts and substantial data, which limits its accessibility for average users.

In contrast, we introduce \textit{prompt-based knowledge distillation}. This novel approach leverages in-context learning (ICL) to facilitate knowledge transfer without the extensive fine-tuning traditionally required. It conditions a low-resource student model on carefully crafted prompts derived from the high-resource teacher model, significantly reducing computational demands and enabling rapid adaptation to new tasks.

Consider the general framework for prompt-based knowledge distillation, where a teacher model $\mathcal{T}$ and a student model $\mathcal{S}$ interact. The teacher model processes an input question $q$ to generate an informative prompt $P$, which encapsulates key insights or directions rather than explicit answers:
\[
\mathcal{T}(q) \rightarrow P
\]
The student model $\mathcal{S}$ then uses this prompt to infer the answer $a$, leveraging the distilled knowledge without direct output replication:
\[
\mathcal{S}(P) \rightarrow a
\]

Consider an educational scenario where a student model is required to solve geometry problems involving circle areas. For the problem "Calculate the area of a circle with a diameter of 10 cm," a high-resource teacher model could generate a prompt that distills essential concepts into several key points:

\begin{itemize}
    \item Remember that the radius is half the diameter.
    \item Use the area formula for a circle: \( \pi r^2 \).
    \item Always include units in your answer (e.g., square cm).
\end{itemize}

This structured prompt guides the student model to focus on the fundamental mathematical relationships and proper problem-solving practices. By applying these principles, the student model calculates the radius as 5 cm and then uses the formula to determine the area as \( 25\pi \) square cm. This approach not only aids in solving the current problem but also reinforces good mathematical practices for future tasks.

\section{Trace-of-Thought Prompting}
Many problems in domains such as arithmetic reasoning can be broken down into intermediate steps that mimic the cognitive process of a human evaluator. Trace-of-Thought Prompting, an application of the prompt-based knowledge distillation framework introduced earlier, enhances models' problem-solving capabilities by breaking down such problems into simpler, actionable steps.

\subsection{Formalization}
We define a general language model $\mathcal{L}$ that processes an input $\mathcal{I}$ into an output $\mathcal{O}$:
\[
\mathcal{L}(\mathcal{I}) \to \mathcal{O}
\]
Assuming our input $q$ is a problem that can be decomposed, we structure it into a sequence of interdependent steps:
\[
q \to \{s_1, s_2, \ldots, s_n\}
\]
The first step in Trace-of-Thought Prompting involves the decomposition of the problem into steps interpretable by a target model. The teacher model, $\mathcal{L}_T$, approximates the set of steps required to solve $q$:
\[
\mathcal{L}_T(q) \stackrel{\approx}{\to} \{s_1, s_2, \ldots, s_n\}
\]
These steps are then used by the student model, $\mathcal{L}_S$, which is tasked with solving the original problem conditioned on the provided steps, aiming to generate the correct answer $a$:
\[
\mathcal{L}_S(q | \{s\}) \to a
\]

\begin{table*}[t]
    \centering
    \small
    \begin{adjustbox}{max width=\textwidth}
        \begin{tabular}{@{}lcccccc@{}}
        \toprule
        \textbf{Model Name} & \textbf{Standard} & \textbf{Chain-of-Thought} & \textbf{Plan \& Solve} & \textbf{Trace-of-Thought (GPT-4)} & \textbf{Trace-of-Thought (Llama 3)}\\
        \midrule
        \rowcolor{gray!20} \multicolumn{6}{c}{\textbf{GSM8K} ($n=200$)} \\
        GPT-4 & 94.5 & \textbf{\cellcolor{lightgreen!30}95.5} & \textbf{\cellcolor{lightgreen!30}95.5} & 95 & 83\\
        GPT-3.5-Turbo & 75.5 & 73.5 & 74.5 & \textbf{\cellcolor{lightgreen!30}86.5\textsuperscript{$\alpha$}} & 64.5\\
        WizardMath-7B & 69 & 73.5 & \textbf{\cellcolor{lightgreen!30}82.5} & 81.5 & 70.5\\
        Llama 3 Chat 8B & 73 & 73 & 68.5 & \textbf{\cellcolor{lightgreen!30}88\textsuperscript{$\alpha$}} & 63.5\\
        Llama 2 Chat 7B & 22 & 23.5 & 23 & \textbf{\cellcolor{lightgreen!30}50\textsuperscript{$\alpha$}} & 37.5\textsuperscript{$\alpha$}\\
        Zephyr & 26 & 23.5 & 30 & \textbf{\cellcolor{lightgreen!30}55\textsuperscript{$\alpha$}} & 43\\
        \rowcolor{gray!20} \multicolumn{6}{c}{\textbf{MATH} ($n=200$)} \\
        GPT-4 & 57.5 & 66 & \textbf{\cellcolor{lightgreen!30}75} & 68 & 55\\
        GPT-3.5-Turbo & 46.5 & 52 & \textbf{\cellcolor{lightgreen!30}56} & \textbf{\cellcolor{lightgreen!30}56} & 40.5\\
        WizardMath-7B & \textbf{\cellcolor{lightgreen!30}44.5} & 33.5 & 37 & 42.5 & 30.5\\
        Llama 3 Chat 8B & 30.5 & 35.5 & 30 & \textbf{\cellcolor{lightgreen!30}41} & 23\\
        Llama 2 Chat 7B & 6.5 & 7 & 5 & \textbf{\cellcolor{lightgreen!30}8} & 6.5\\
        Zephyr & 7 & 12 & 9 & 13.5 & \textbf{\cellcolor{lightgreen!30}14.5}\\
        \bottomrule
        \end{tabular}
    \end{adjustbox}
        \caption{
        Evaluation results for both GSM8K and MATH, $n=200$. $\alpha$ denotes results where Trace-of-Thought's gain over the highest alternative was significant at $\alpha=0.05$ (see Tables \ref{tab:siggsm8khigh}, \ref{tab:sigmathhigh}, \ref{tab:siggsm8klow} and \ref{tab:sigmathlow})}
    \label{tab:results}
\end{table*}

\subsection{Practical Application Example}
Consider the following GSM8K problem \( P \): "Natalia sold clips to 48 of her friends in April, and then she sold half as many clips in May. How many clips did Natalia sell altogether in April and May?"

\textbf{Teacher Model – Delegation Phase:}
\begin{quote}
\texttt{Create very short step-by-step prompts for the following problem: \( P \)\\
Format as a list. Do not solve the problem.}
\end{quote}

The teacher model might generate these steps:
\begin{itemize}
    \item Identify April's sales.
    \item Calculate May's sales as half of April's.
    \item Add April's and May's sales to find the total.
\end{itemize}

\textbf{Student Model – Solution Phase:}
\begin{quote}
\texttt{We are given the following problem: \( P \)\\
Use the following steps to solve the problem:\\
1. April sales: 48 clips.\\
2. May sales: 24 clips.\\
3. Total sales: 48 + 24 = 72 clips.}
\end{quote}

The student model uses the steps provided to compute the final answer: 72 clips. This approach not only ensures the student model understands the process of solving the problem but also maintains the structure of the reasoning path laid out by the teacher model.

Table \ref{tab:prompt_templates} showcases the exact text necessary for the delegation and solution prompts, where the question and steps are interpolated as needed. A visual comparison with popular prompting approaches is provided in Figure \ref{fig:mainfig}.

\section{Experimental Setup}
\subsection{Benchmarks}
To evaluate the effectiveness of Trace-of-Thought in a practical environment, we select two arithmetic reasoning datasets of varying difficulty:
\begin{enumerate}
    \item \textbf{GSM8K} \citep{cobbe2021training} --- GSM8K is a dataset of 8 thousand grade school level arithmetic reasoning problems, with a focus on simple problems that require some level of variable identification and decomposed reasoning.
    \item \textbf{MATH} \citep{li2023camel} --- MATH is a dataset of 50 thousand synthetically generated mathematical reasoning problems; MATH primarily focuses on a mix of simple and difficult arithmetic reasoning problems, with extended domains such as complex numbers, geometric reasoning, and functions.
\end{enumerate}
In order to appropriately evaluate performance on these datasets, we sample $n=200$ examples from each dataset, using each of the prompts in Table \ref{tab:prompt_templates} on a suite of models.
\subsection{Prompting Approaches}
To evaluate each sampled problem, we employ a suite of popular prompting approaches in the literature:
\begin{enumerate}
    \item \textbf{Zero-Shot Standard Prompting} --- where each sampled question makes up the sole input to the model, with no in-context information provided.
    \item \textbf{Zero-Shot Chain-of-Thought Prompting} --- where each sampled question is appended with instructions to "think step-by-step" as proposed in \citet{wei2023chainofthought} and \citet{kojima2023large}.
    \item \textbf{Zero-Shot Plan \& Solve} --- where models are instructed to process the question, devise a plan, and solve that plan step-by-step prior to the question being provided as proposed in \citet{wang2023planandsolve}.
    \item \textbf{Zero-Shot Trace-of-Thought Prompting} --- where a model is first instructed to decompose a problem into steps, before those steps are passed to another model instance for solution. Two variants are employed: \textbf{GPT-4} as a teacher model, investigating high- to low-resource distillation; and \textbf{Llama 3 8B} as a teacher model, investigating low- to low-resource distillation.\footnote{Note that while Tree of Thoughts and Least-to-Most Prompting also fall under decomposition frameworks, their recursive nature is often difficult to properly emulate and does not align with the linear approaches suggested herein.}
    
\end{enumerate}

\subsection{Evaluation}
After a question is fully solved, the inputs, outputs, and provided dataset output are written to a file for human evaluation. The full set of testing data, comprised of 12 thousand total samples, is then human annotated by the authors, collectively familiar with all mathematical concepts leveraged by either dataset; answers are annotated with a 1 if the outputs are in line with the labels provided, and a 0 otherwise. The resulting score, given out of 200, is then tabulated as a percentage accuracy score for reporting.\footnote{The data files used for evaluation, along with the scripts for analysis, will be made available in a public repository linked in the Abstract. Comprehensive documentation will accompany the data to assist researchers in replicating and extending the study.}

\section{Results}
Table \ref{tab:results} reports the accuracy results of each model and prompting approach on both datasets; the uppermost table corresponds to results on GSM8K, while the lower table corresponds to results for MATH.
\subsection{High-Resource Teachers - GPT-4}
When applying GPT-4 as a high-resource teacher, on 58.3\% of testing suites, high-resource Trace-of-Thought generates results with the highest accuracy scores across both GSM8K and MATH. While some gains are slightly more nuanced --- such as those observed when applied to GPT-4 on MATH --- many low-resource models see strong accuracy gains when endowed with critical reasoning distilled from GPT-4. In the greatest of such cases, Llama 2's performance on GSM8K sees a rise of 27\% absolute accuracy from 23\% to 50\% when queried using Trace-of-Thought.

\subsection{Low-Resource Teachers - Llama 3}
While Llama 3 as a teacher model does not encourage such gains as GPT-4, we observe that traditionally less performant models such as Llama 2 and Zephyr benefit strongly from distillation from a much smaller model than that of a high-resource teacher. On GSM8K, and with just a 14\% size difference between teacher and student, we observe absolute accuracy gains of 14.5\% and 13\% on Llama 2 and Zephyr respectively. 

\subsection{Relative Accuracy Changes}
\begin{table}[t]
    \centering
    \small
    \begin{adjustbox}{max width=\columnwidth}
        \begin{tabular}{@{}lcccccc@{}}
        \toprule
        \textbf{Model} & $\bar{x}_{HPA}$ & \textbf{Trace-of-Thought} & \textbf{\% Gain} \\
        \midrule
        \rowcolor{gray!20} \multicolumn{4}{c}{\textbf{GSM8K} ($n=200$)} \\
        GPT-4 & 95.5 & 95 & -0.52 \\
        GPT-3.5-Turbo & 75.5 & 86.5 & \textbf{\cellcolor{lightgreen!30}14.57} \\
        Llama 3 8B & 73 & 88 & \textbf{\cellcolor{lightgreen!30}20.55} \\
        WizardMath-7B & 82.5 & 81.5 & -1.21 \\
        Llama 2 7B & 23.5 & 50 & \textbf{\cellcolor{lightgreen!30}112.77} \\
        Zephyr-7B & 30 & 55 & \textbf{\cellcolor{lightgreen!30}83.3} \\
        \midrule
        \rowcolor{gray!20} \multicolumn{4}{c}{\textbf{MATH} ($n=200$)} \\
        GPT-4 & 75 & 68 & \textbf{\cellcolor{lightgreen!30}3.03} \\
        GPT-3.5-Turbo & 56 & 56 & 0\\
        Llama 3 8B & 35.5 & 41 & \textbf{\cellcolor{lightgreen!30}15.49} \\
        WizardMath-7B & 44.5 & 42.5 & -4.49 \\
        Llama 2 7B & 7 & 8 & \textbf{\cellcolor{lightgreen!30}14.29} \\
        Zephyr-7B & 12 & 13.5 & \textbf{\cellcolor{lightgreen!30}12.5} \\
        \bottomrule
        \end{tabular}
    \end{adjustbox}
    \caption{Relative gain over highest performing alternative approach ($\bar{x}_{HPA}$) - \textbf{high-resource teacher} (GPT-4).}
    \label{tab:relative_high}
\end{table}

\begin{table}[t]
    \centering
    \small
    \begin{adjustbox}{max width=\columnwidth}
        \begin{tabular}{@{}lcccccc@{}}
        \toprule
        \textbf{Model} & $\bar{x}_{HPA}$ & \textbf{Trace-of-Thought} & \textbf{\% Gain} \\
        \midrule
        \rowcolor{gray!20} \multicolumn{4}{c}{\textbf{GSM8K} ($n=200$)} \\
        GPT-4 & 95.5 & 83 & -13.09 \\
        GPT-3.5-Turbo & 75.5 & 64.5 & -14.57 \\
        Llama 3 8B & 73 & 63.5 & -13.01 \\
        WizardMath-7B & 82.5 & 70.5 & -14.55 \\
        Llama 2 7B & 23.5 & 37.5 & \textbf{\cellcolor{lightgreen!30}59.57} \\
        Zephyr-7B & 30 & 43 & \textbf{\cellcolor{lightgreen!30}43.33} \\
        \midrule
        \rowcolor{gray!20} \multicolumn{4}{c}{\textbf{MATH} ($n=200$)} \\
        GPT-4 & 75 & 55 & -26.67 \\
        GPT-3.5-Turbo & 56 & 40.5 & -27.68 \\
        Llama 3 8B & 35.5 & 23 & -35.21 \\
        WizardMath-7B & 44.5 & 30.5 & -31.46 \\
        Llama 2 7B & 7 & 6.5 & -7.14 \\
        Zephyr-7B & 12 & 14.5 & \textbf{\cellcolor{lightgreen!30}20.83} \\
        \bottomrule
        \end{tabular}
    \end{adjustbox}
    \caption{Relative gain over highest performing alternative approach ($\bar{x}_{HPA}$) - \textbf{low-resource teacher} (Llama 3).}
    \label{tab:relative_low}
\end{table}
As mentioned, there is an inherent issue of scale when considering performance improvements or drawbacks of using Trace-of-Thought. Tables \ref{tab:relative_high} and \ref{tab:relative_low} show the relative gains or losses of Trace-of-Thought on each student model at both teacher model scales.

As previously stated, a majority of models benefit from high-resource distillation with GPT-4; gains tend to be slightly more incremental on other higher-resource models (GPT-4, GPT-3.5-Turbo) or domain fine-tuned (WizardMath-7B) models, while gains are more notable on models of less scale and ability, exceeding 100\%. 

\section{Distillation Ability vs. Absolute Performance}
\begin{figure}[t]
    \centering
        \includegraphics[width=\columnwidth, keepaspectratio]{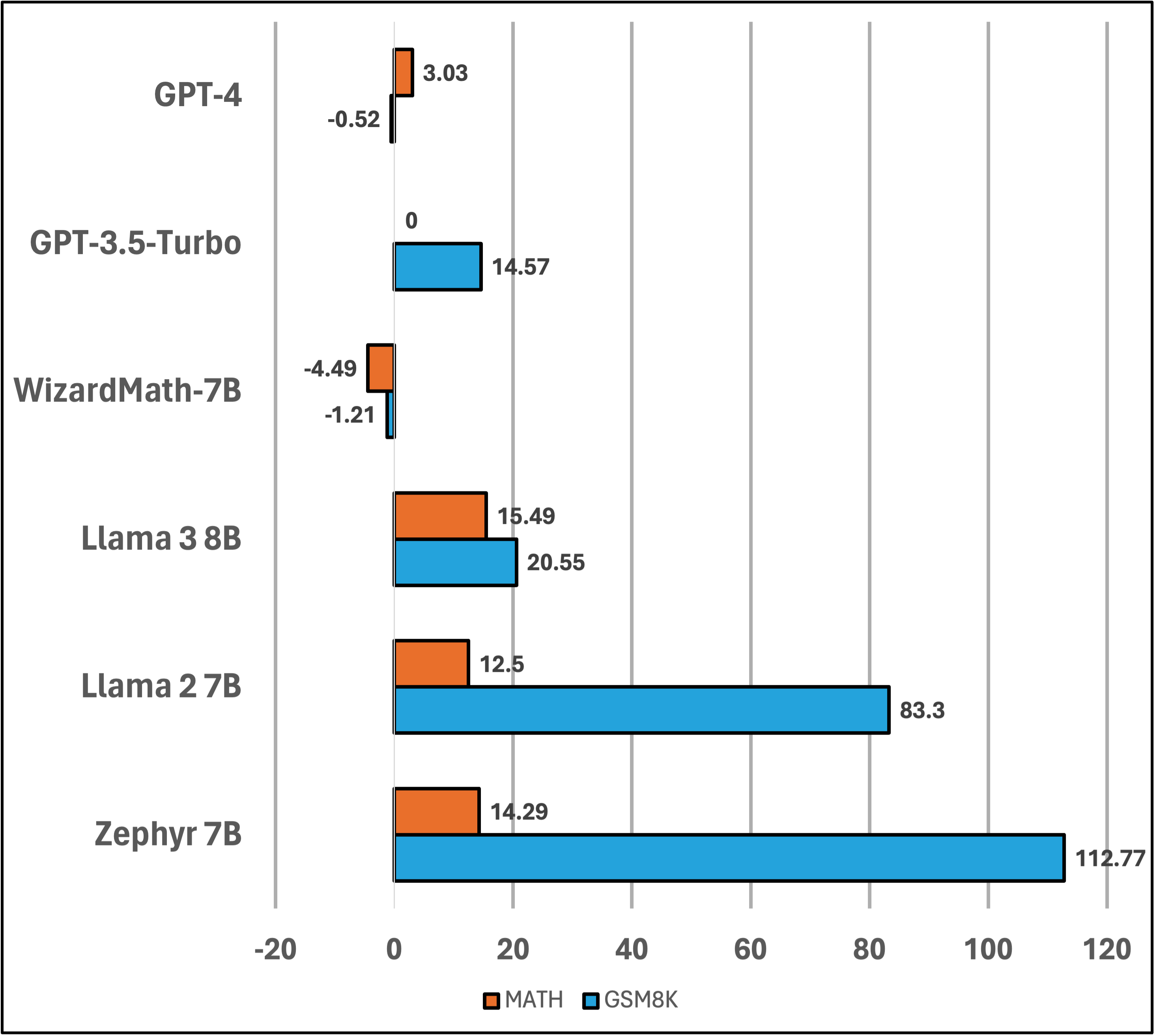}
    \caption{Relative accuracy changes with Trace-of-Thought (high-resource) visualized, in order of absolute performance.}
    \label{fig:hrabsolutechange}
\end{figure}

\begin{figure}[t]
    \centering
        \includegraphics[width=\columnwidth, keepaspectratio]{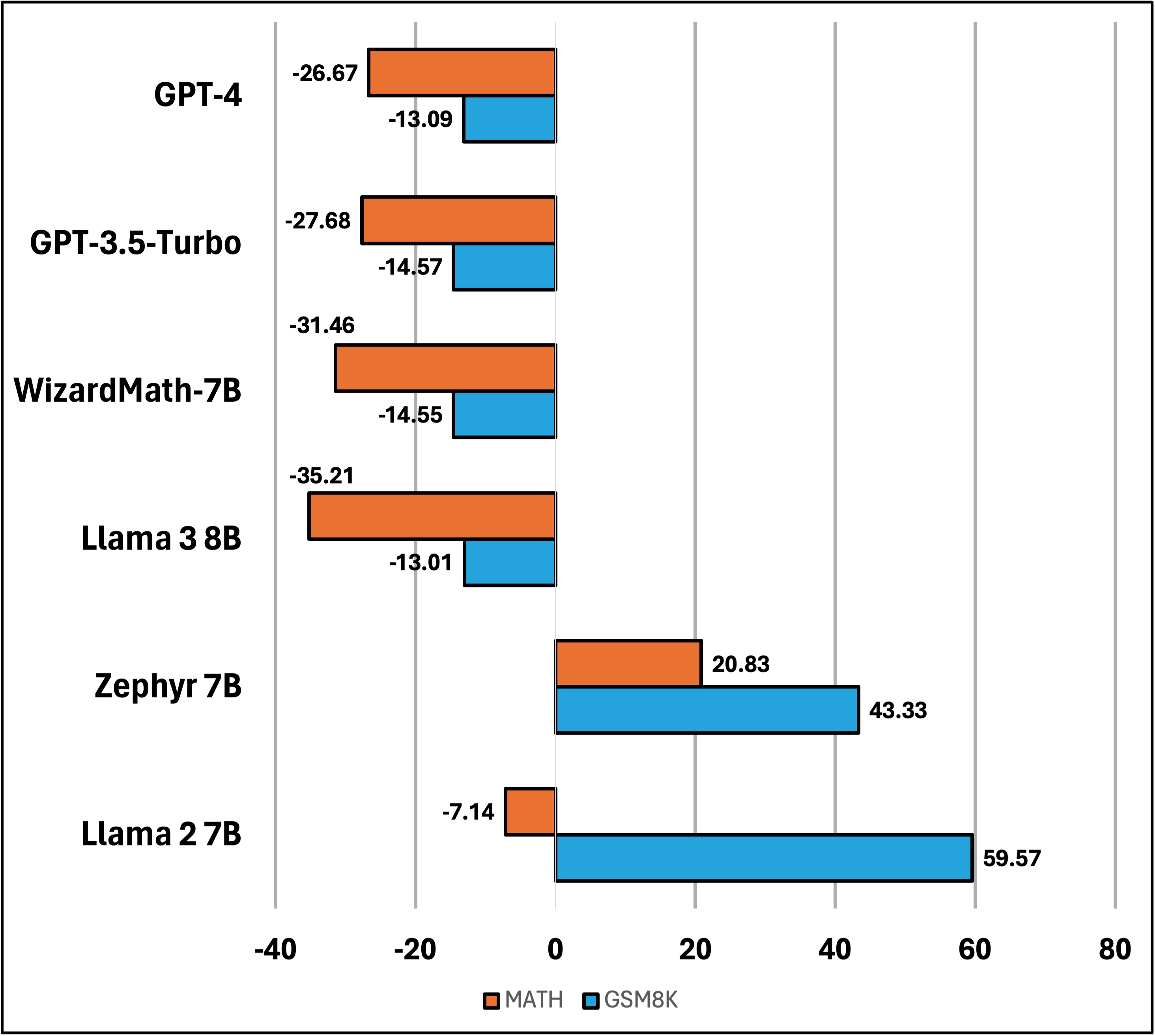}
    \caption{Relative accuracy changes with Trace-of-Thought (low-resource) visualized, in order of absolute performance.}
    \label{fig:lrabsolutechange}
\end{figure}

Figures \ref{fig:hrabsolutechange} and \ref{fig:lrabsolutechange} report relative gains sorted by average of absolute performance, or the average of a model's performance on every approach for each dataset. 

High-resource teachers are shown to have variable levels of correlation ($r=0.8363$ on GSM8K; $r=0.4703$ on MATH) between a model's absolute ability and the performance of Trace-of-Thought on said model; this trend also holds for low-resource models, who tend to succeed with distilling to historically under-performant models ($r=0.7886$ on GSM8K; $r=0.7595$ on MATH).

By applying further strategies, such as fine-tuning, there is great potential for low-resource teachers to be easily deployable critical reasoners, without the financial or customization burdens of using closed-source models. However --- even innately --- our results show that there is already exceedingly beneficial use-cases for low-resource applications of distillation.
\section{Qualitative Analysis}


\subsection{Transparent Reasoning}
Trace-of-Thought Prompting enhances transparency in reasoning processes compared to previous approaches, allowing for more effective human-in-the-loop or automated corrections before erroneous reasoning is executed. This capability is crucial, as highlighted by \citet{wu2024mitigating}, who discuss the challenges of interpretability and faithfulness in models that only allow post-hoc evaluations. For instance, consider a problem from the GSM8K dataset:

\begin{displayquote}
\textit{Input: If Marcy works for the same company for 40 years, she gets an annual pension of \$50,000/year. Starting after 20 years, she becomes entitled to 5\% of the value of the pension per year. If she quits after 30 years, what will her annual pension be?}\\
Output Steps:
\begin{itemize}
    \item Determine how many years Marcy worked before becoming entitled to additional pension benefits (20 years).
    \item \textbf{*Calculate the value of Marcy's annual pension after 20 years. This is her basic pension without the additional benefit.} (erroneous)
\end{itemize}
\end{displayquote}

In this GSM8K example, the Trace-of-Thought method allows for identifying and correcting the misinterpretation of the pension calculation rules before finalizing the answer, which is an advantage in educational and practical applications.

\subsection{Adaptive Reasoning}
Trace-of-Thought enhances the adaptive reasoning capabilities of models, allowing them to adjust their problem-solving strategies based on the provided prompts. This adaptability reflects human cognitive processes and is vital for handling complex, real-world scenarios. \citet{didolkar2024metacognitive} highlights how this flexibility enhances models' ability to generalize and creatively apply knowledge. An example from the GSM8K dataset demonstrates this:

\begin{displayquote}
    \textit{Input: Consider the function $g(x) = 3$. Find $g(2)$.}
    Steps:
    \begin{itemize}
        \item Review the function definition: **g(x) = 3**
        \item Identify the target: **g(2)**
        \item Substitute x with 2 in the function: **g(2) = ?**
        \item Acknowledge the function's constant value: **? = 3**
    \end{itemize}
    \textbf{Output: Given $g(x) = 3$, substituting any value for x results in 3. Thus, $g(2) = 3$.}
\end{displayquote}

In this case, the model efficiently applies its understanding that the output of a constant function is unchanging, thereby eliminating unnecessary computational steps. Moreover, Trace-of-Thought's flexible framework supports a form of interactive learning where models can both explore new problem-solving pathways and refine existing strategies. This interactive and adaptive approach mirrors the dynamics of human learning and collaboration, suggesting further avenues for research into the cooperative aspects of AI reasoning as noted by \citet{chen2024comm}.

\begin{table}[t]
    \centering
    \small
    \begin{adjustbox}{max width=\columnwidth}
        \begin{tabular}{@{}p{0.3\columnwidth}p{0.35\columnwidth}p{0.3\columnwidth}@{}}
        \toprule
        \textbf{Error Type} & \textbf{Input} & \textbf{Output}\\
        \midrule
        Llama 3 - Incorrect Distillation & Carla is downloading a 200 GB file. Normally she can download 2 GB/minute, but 40\% of the way through the download, Windows forces a restart to install updates, which takes 20 minutes. \textbf{Then Carla has to restart the download from the beginning.} How load does it take to download the file? & *Step 6: Calculate the new download time
Calculate how long it will take Carla to \textbf{download the remaining 60\%} of the file at her usual rate.\newline \newline \textbf{Result (Zephyr): *260 mins.} \newline (Correct: 180)\\
        \midrule
        GPT-4 - Incorrect Distillation & If Marcy works for the same company for 40 years, she gets an annual pension of \$50,000/year. \textbf{Starting after 20 years, she becomes entitled to 5\% of the value of the pension per year.} If she quits after 30 years, what will her annual pension be? & *3. \textbf{Calculate the base pension that Marcy is eligible for after 20 years.} This is because after 20 years, she starts accumulating an additional 5\% per year. \newline \newline \textbf{Result (Llama 2): *\$378,125} \newline (Correct: \$25,000)\\
        \bottomrule
        \end{tabular}
    \end{adjustbox}
    \caption{Teacher errors encountered when employing GPT-4 or Llama 3 as a teacher model.}
    \label{tab:teacherref}
\end{table}

\subsection{Error Analysis}
Table \ref{tab:teacherref} illustrates examples of teacher errors, or errors in the distillation process; table (ref) subsequently demonstrates examples of student errors, or traditional reasoning flaws.

While traditional faults in reasoning can be observed from student models, there are instances where distilled reasoning can be altered by an erroneous distillation from the teacher model. As described in the table, arithmetic inconsistencies can propagate from the teacher to the student if the self-correction abilities of the student are not strong.

Similarly, we observe that a strong distillation can occasionally result in a poor solution; when processing distillations that contain higher-level steps, models with poor self-correction abilities also tend to adhere less strictly to distilled instructions.

\begin{table}[t]
    \centering
    \small
    \begin{adjustbox}{max width=\columnwidth}
        \begin{tabular}{@{}p{0.3\columnwidth}p{0.35\columnwidth}p{0.3\columnwidth}@{}}
        \toprule
        \textbf{Error Type} & \textbf{Input \& Steps} & \textbf{Output}\\
        \midrule
        Zephyr - Incorrect Solution & James decides to run 3 sprints 3 times a week.  He runs 60 meters each sprint.  How many total meters does he run a week?\newline...\newline4. Identify the number of days James runs in a week. \newline 5. \textbf{Multiply the total meters run each day by the number of days he runs each week to find the total meters run in a week.} & 5. To find the total meters James runs in a week, we multiply the total meters he runs each day by the number of days he runs: 180 meters per day x 9 days = \textbf{*1620 meters in one week.}\newline(Correct: 540m)\\
        \midrule
        Llama 2 - Incorrect Solution & A candle melts by 2 centimeters every hour that it burns. How many centimeters shorter will a candle be after burning from 1:00 PM to 5:00 PM?\newline ...\newline5. To find out how many centimeters shorter the candle will be, \textbf{multiply the total burning hours by the melt rate.} & 5. To find out how many centimeters shorter the candle will be, multiply the total burning hours by the melt rate: 5 hours x 2 cm/hour = 10 cm. So, after burning from 1:00 PM to 5:00 PM, \textbf{*the candle will be 10 centimeters shorter.}\newline (Correct: 8cm)\\
        \bottomrule
        \end{tabular}
    \end{adjustbox}
    \caption{Student errors encountered when employing Zephyr or Llama 2 as a student model.}
    \label{tab:studentref}
\end{table}

\section{Conclusion}
This paper introduces a structured approach to prompt-based knowledge distillation, building on traditional methods to enhance accessibility and practicality for end-users. Our methodology, Trace-of-Thought Prompting, serves as a practical implementation of this framework, designed to facilitate problem decomposition and improve problem-solving capabilities in both high-resource and low-resource models. Through our experiments with different teacher model sizes, we have demonstrated how Trace-of-Thought can effectively leverage the knowledge distilled from both large and small models, improving reasoning capabilities in a variety of contexts. Our results show significant gains in model performance, especially in scenarios involving low-resource models, highlighting the potential of this approach to make AI more accessible and effective for a broader range of applications.
\section*{Limitations}

\textbf{Abstract reasoning.} Datasets such as ARC and ACRE explore abstract reasoning tasks, where the aim is to generalize and apply a common pattern across very little data \cite{xu2024llms, gendron2024large}. Due to their abstract nature, and the lack of observable divisions of the original problem, it remains to be seen whether Trace-of-Thought is a valid and useful framework for conducting tasks such as general pattern recognition or commonsense reasoning.

\textbf{Computational impact.} Due to the extra prompt necessary for generating steps for the solution model, and the lengthier input prompts necessary to achieve this effect, applying Trace-of-Thought in a primarily closed-source setting may lead to additional costs and heightened usage of restricted resources such as API credits toward rate limits.

\textbf{Solution diffusion from delegation model.} It remains to be fully investigated whether the decomposition prompt in the first step may be too strong on tasks with complex reasoning; too strong of a prompt may encourage the teacher model to solve the problem in order to confirm its own reasoning, leading to inadvertent data contamination.


\bibliography{custom}
\clearpage
\onecolumn
\section*{Appendix}
\begin{table}[h]
\centering
\begin{adjustbox}{width=\textwidth}
\begin{tabular}{|l|r|r|r|r|}
\multicolumn{5}{c}{Two Sample Z-Test for Proportions - significant gains are bolded, and their significance level is put in brackets.}\\
  \hline
Model & Trace-of-Thought($\Bar{x}_{ToT}$) & Highest Performing Alternative ($\Bar{x}_{HPA}$) & $z$ & $p$ \\
\hline
GPT-4 & 95 & 95.5 & -0.1662 & ---\\
GPT-3.5-Turbo & 86.5 & 75.5 & 1.9827 & \textbf{0.0477 ($p<0.05$)}\\
Llama 3 8B & 88 & 73 & 2.6771 & \textbf{0.00736 ($p<0.01$)}\\
WizardMath-7B & 81.5 & 82.5 & -0.1841 & ---\\
Llama 2 7B Chat & 50 & 23.5 & 3.8866 & \textbf{0.0001 ($p<0.01$)}\\
Zephyr-7B & 55 & 30 & 3.576 & \textbf{0.00034 ($p<0.01$)}\\
\hline
\end{tabular}
\end{adjustbox}
\caption{Comparison of \textbf{high-resource} Trace-of-Thought performance against highest performing alternatives on the GSM8K dataset using two sample Z-test for proportions, $\alpha=0.05$. Only scenarios with positive Z (gains) are reported.}
\label{tab:siggsm8khigh}
\end{table}

\begin{table}[htp!]

\centering
\begin{adjustbox}{width=\textwidth}
\begin{tabular}{|l|r|r|r|r|}
\multicolumn{5}{c}{Two Sample Z-Test for Proportions - significant gains are bolded, and their significance level is put in brackets.}\\
  \hline
Model & Trace-of-Thought($\Bar{x}_{ToT}$) & Highest Performing Alternative ($\Bar{x}_{HPA}$) & $z$ & $p$ \\
\hline
GPT-4 & 68 & 75 & -1.0965 & ---\\
GPT-3.5-Turbo & 56 & 56 & 0 & ---\\
Llama 3 8B & 41 & 35.5 & 0.8002 & 0.42372\\
WizardMath-7B & 42.5 & 44.5 & -0.2853 & ---\\
Llama 2-7B Chat & 8 & 7 & 0.2685 & 0.78716\\
Zephyr-7B & 13.5 & 12 & 0.318 & 0.74896\\
\hline
\end{tabular}
\end{adjustbox}
\caption{Comparison of \textbf{high-resource} Trace-of-Thought performance against highest performing alternatives on the MATH dataset using two sample Z-test for proportions, $\alpha=0.05$. Only scenarios with positive Z (gains) are reported.}
\label{tab:sigmathhigh}
\end{table}

\begin{table}[h]
\centering
\begin{adjustbox}{width=\textwidth}
\begin{tabular}{|l|r|r|r|r|}
\multicolumn{5}{c}{Two Sample Z-Test for Proportions - significant gains are bolded, and their significance level is put in brackets.}\\
  \hline
Model & Trace-of-Thought($\Bar{x}_{ToT}$) & Highest Performing Alternative ($\Bar{x}_{HPA}$) & $z$ & $p$ \\
\hline
GPT-4 & 83 & 95.5 & -2.8536 & ---\\
GPT-3.5-Turbo & 64.5 & 75.5 & -1.6973 & ---\\
Llama 3 8B & 63.5 & 73 & -1.4431 & ---\\
WizardMath-7B & 70.5 & 82.5 & -2.0013 & ---\\
Llama 2 7B Chat & 37.5 & 23.5 & 2.1502 & \textbf{0.03156 ($p<0.05$)}\\
Zephyr-7B & 43 & 30 & 1.9094 & 0.05614\\
\hline
\end{tabular}
\end{adjustbox}
\caption{Comparison of \textbf{low-resource} Trace-of-Thought performance against highest performing alternatives on the GSM8K dataset using two sample Z-test for proportions, $\alpha=0.05$. Only scenarios with positive Z (gains) are reported.}
\label{tab:siggsm8klow}
\end{table}

\begin{table}[htp!]

\centering
\begin{adjustbox}{width=\textwidth}
\begin{tabular}{|l|r|r|r|r|}
\multicolumn{5}{c}{Two Sample Z-Test for Proportions - significant gains are bolded, and their significance level is put in brackets.}\\
  \hline
Model & Trace-of-Thought($\Bar{x}_{ToT}$) & Highest Performing Alternative ($\Bar{x}_{HPA}$) & $z$ & $p$ \\
\hline
GPT-4 & 55 & 75 & -2.965 & ---\\
GPT-3.5-Turbo & 40.5 & 56 & -2.1934 & ---\\
Llama 3 8B & 23 & 35.5 & -1.943 & ---\\
WizardMath-7B & 30.5 & 44.5 & -2.0448 & ---\\
Llama 2-7B Chat & 6.5 & 7 & -0.1409 & ---\\
Zephyr-7B & 14.5 & 12 & 0.5214 & 0.60306\\
\hline
\end{tabular}
\end{adjustbox}
\caption{Comparison of \textbf{low-resource} Trace-of-Thought performance against highest performing alternatives on the MATH dataset using two sample Z-test for proportions, $\alpha=0.05$. Only scenarios with positive Z (gains) are reported.}
\label{tab:sigmathlow}
\end{table}

\end{document}